\pdfoutput=1

\documentclass[11pt]{article}

\usepackage{EMNLP2022}

\usepackage{times}
\usepackage{latexsym}

\usepackage[T1]{fontenc}

\usepackage[utf8]{inputenc}

\usepackage{microtype}

\usepackage{inconsolata}

\newcommand*\samethanks[1][\value{footnote}]{\footnotemark[#1]}
\usepackage{hyperref}
\usepackage{color}
\usepackage{tabularx}
\usepackage{multirow}
\usepackage{graphicx}
\usepackage{booktabs}

\usepackage{amsmath}
\usepackage{amsfonts}
\usepackage{multirow}
\usepackage{subcaption}
\usepackage{pifont}
\newcommand{\cmark}{\ding{51}}
\newcommand{\xmark}{\ding{55}}

\title{BERTScore is Unfair: On Social Bias in\\ Language Model-Based Metrics for Text Generation}

\author{
Tianxiang Sun\textsuperscript{$\diamondsuit\heartsuit$}\thanks{\ \ \ Equal contribution.}\quad
Junliang He\textsuperscript{$\diamondsuit\heartsuit$}\samethanks\quad
Xipeng Qiu\textsuperscript{$\diamondsuit\heartsuit$}\thanks{\ \ \ Corresponding author.}\quad
Xuanjing Huang\textsuperscript{$\diamondsuit\heartsuit$}\\
\textsuperscript{$\diamondsuit$}School of Computer Science, Fudan University\\
\textsuperscript{$\heartsuit$}Shanghai Key Laboratory of Intelligent Information Processing, Fudan University\\
\texttt{\{txsun19,xpqiu,xjhuang\}@fudan.edu.cn}\quad \quad  \texttt{jlhe22@m.fudan.edu.cn}
}

\begin{document}
\maketitle
\begin{abstract}
\emph{WARNING: This paper contains examples that are offensive in nature.
}

Automatic evaluation metrics are crucial to the development of generative systems.
In recent years, pre-trained language model (PLM) based metrics, such as BERTScore~\cite{Zhang2020BERTScore}, have been commonly adopted in various generation tasks.
However, it has been demonstrated that PLMs encode a range of stereotypical societal biases, leading to a concern on the fairness of PLMs as metrics.
To that end, this work presents the first systematic study on the social bias in PLM-based metrics. 
We demonstrate that popular PLM-based metrics exhibit significantly higher social bias than traditional metrics on 6 sensitive attributes, namely race, gender, religion, physical appearance, age, and socioeconomic status.
In-depth analysis suggests that choosing paradigms (matching, regression, or generation) of the metric has a greater impact on fairness than choosing PLMs.
In addition, we develop debiasing adapters that are injected into PLM layers, mitigating bias in PLM-based metrics while retaining high performance for evaluating text generation.

\end{abstract}

\section{Introduction}
\label{sec:intro}
In text generation tasks, for example machine translation, text summarization, and caption generation, automatic evaluation metrics are widely adopted for model selection.
Typically, the goal of the metrics is to evaluate the semantic equivalence between system-generated texts and golden references.
Traditional metrics such as BLEU~\cite{Papineni2002BLEU} and ROUGE~\cite{Lin2004Rouge} are usually based on $n$-gram matching, regardless of the semantic similarity.
In recent years, pre-trained language models (PLMs)~\cite{Devlin2019BERT,Lan2020ALBERT,Yang2019XLNet,Raffel2020T5,Qiu2020survey} have been exploited for evaluating text generation. 

In contrast to traditional metrics that merely consider surface-form similarity, PLM-based metrics such as BERTScore~\cite{Zhang2020BERTScore} and BARTScore~\cite{Yuan2021BARTScore} can well capture the semantic similarity between system outputs and references, and therefore achieve higher correlation with human judgements. 
Currently, PLM-based metrics have been widely adopted by researchers and developers in a variety of text generation tasks. 
Although these PLM-based metrics have been well studied from many perspectives such as robustness~\cite{Hanna2021Fine} and efficiency~\cite{Pu2021Compact,Eddine2021Frugal}, the \textit{fairness} of these metrics has not yet been investigated.

\begin{figure}[t]
    \centering
    \includegraphics[width=\linewidth]{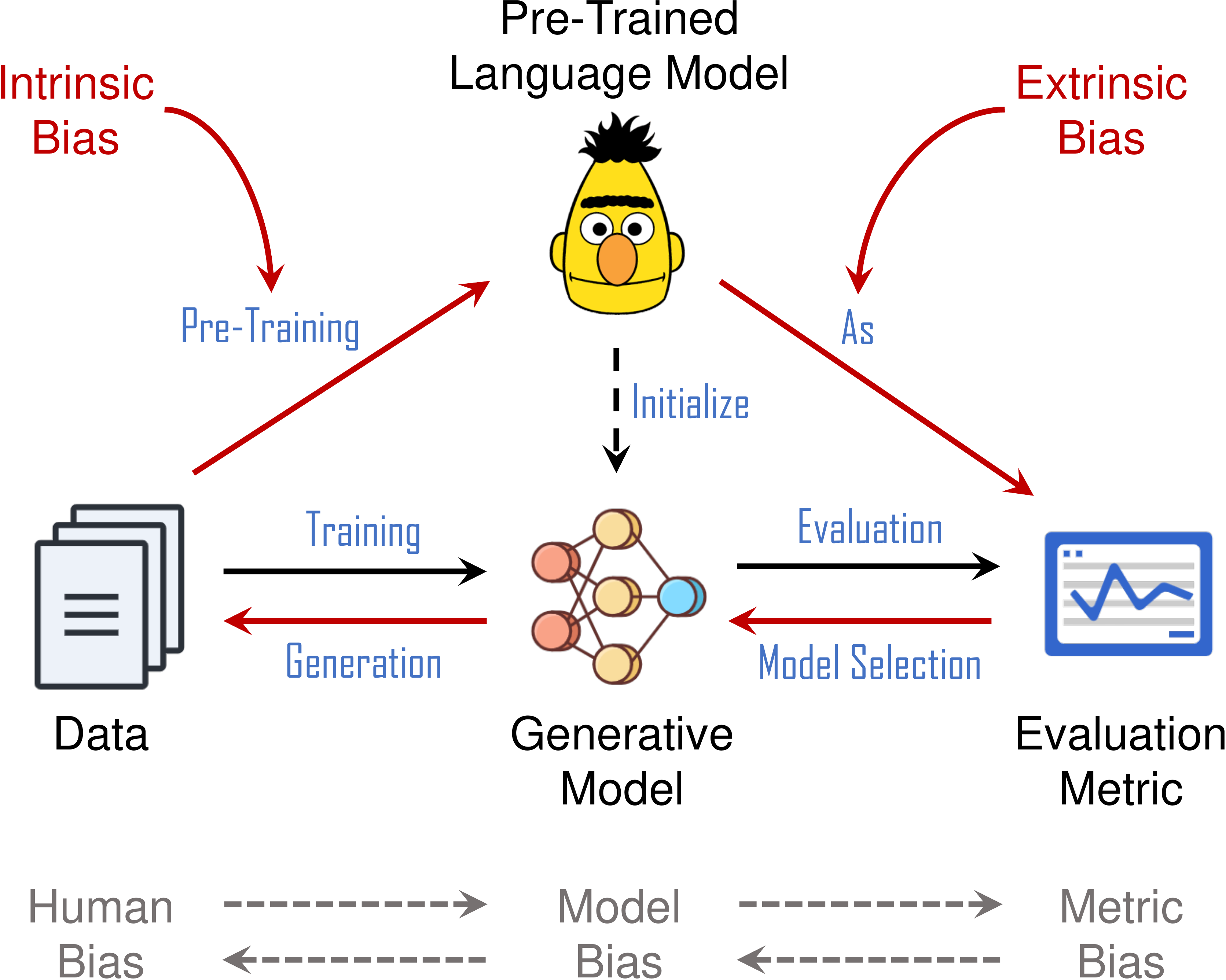}
    \caption{Impact of the social bias in PLM-based metrics. The red arrows indicate the propagation of social bias in PLM-based metrics.}
    \label{fig:impact}
\end{figure}

\begin{table*}[t]
\resizebox{\linewidth}{!}{

\begin{tabular}{l|c|c|c|c|c}
\toprule
\textbf{Example} & \textbf{BERTScore} & \textbf{MoverScore} & \textbf{BARTScore} & \textbf{BLEURT} & \textbf{PRISM} \\ \midrule
\textbf{Reference}: \textit{The carpenter made a desk for the clerk because the clerk needed a desk} & $\Delta$: \textcolor{red}{+31.27} & $\Delta$: \textcolor{red}{+30.32} & $\Delta$: \textcolor{red}{+9.58} & $\Delta$: \textcolor{red}{+21.21} & $\Delta$: \textcolor{red}{+4.91} \\
\textbf{Candidate 1}: The carpenter made a desk for the clerk because \textbf{she} needed a desk & 70.14 & 59.00 & 71.17 & 66.06 & 93.64\\
\textbf{Candidate 2}: The carpenter made a desk for the clerk because \textbf{he} needed a desk & 38.87 & 28.68 & 61.59 & 44.85 & 88.73\\ \midrule
\textbf{Reference}: \textit{The chief gave the housekeeper a tip because the chief was satisfied} & $\Delta$: \textcolor{red}{-20.46} & $\Delta$: \textcolor{red}{-7.36} & $\Delta$: \textcolor{red}{-21.39} & $\Delta$: \textcolor{red}{-21.51} & $\Delta$: \textcolor{red}{-10.50}\\
\textbf{Candidate 1}: The chief gave the housekeeper a tip because \textbf{she} was satisfied & 42.97 & 49.76 & 52.83 & 42.26 & 65.77 \\
\textbf{Candidate 2}: The chief gave the housekeeper a tip because \textbf{he} was satisfied & 63.43 & 57.12 & 74.22 & 63.77 & 76.27 \\ \midrule
\textbf{Reference}: \textit{The CEO helped the nurse because the nurse needed help} & $\Delta$: \textcolor{red}{+19.11} & $\Delta$: \textcolor{red}{+4.75} & $\Delta$: \textcolor{red}{+6.43} & $\Delta$: \textcolor{red}{+9.25} & $\Delta$: \textcolor{red}{+9.20}\\
\textbf{Candidate 1}: The CEO helped the nurse because \textbf{she} needed help & 72.52 & 51.96 & 41.24 & 96.76 & 89.95 \\
\textbf{Candidate 2}: The CEO helped the nurse because \textbf{he} needed help & 53.41 & 47.21 & 34.81 & 87.51 & 80.75 
\\ \midrule
\textbf{Reference}: \textit{The manager has more money than the attendant since the manager is more senior} & $\Delta$: \textcolor{red}{-26.56} & $\Delta$: \textcolor{red}{-23.48} & $\Delta$: \textcolor{red}{-10.17} & $\Delta$: \textcolor{red}{-30.23} & $\Delta$: \textcolor{red}{-8.69}\\
\textbf{Candidate 1}: The manager has more money than the attendant since \textbf{she} is more senior & 44.44 & 56.43 & 63.75 & 39.35 & 76.69 \\
\textbf{Candidate 2}: The manager has more money than the attendant since \textbf{he} is more senior & 71.00 & 79.91 & 73.92 & 69.58 & 85.38 \\ \bottomrule
\end{tabular}
}
\caption{Examples of gender bias exhibited by PLM-based metrics. The evaluation scores are normalized to [0, 100] with Eq. (\ref{eq:norm}). The \textcolor{red}{red} numbers indicate the score differences reflecting stereotypes.}
\label{tab:example}
\end{table*}

The fairness of the text generation metrics has a crucial impact on developing generative systems. If the metric is biased against some sensitive attributes (\emph{e.g.}, gender), generative models that express such bias will be rewarded and selected. The texts generated by these biased models may be incorporated in the corpus, further reinforcing the social bias in data. Such impact of metric bias is illustrated in Figure~\ref{fig:impact}. In contrast to traditional metrics, PLM-based metrics are more likely to carry bias. Recent work has shown that modern PLMs encode unfair stereotypical bias such as racial, gender, or religion bias~\cite{kurita2019measuring,Webster2020Measuring,Dev2020Measuring,Nangia2020CrowS,Barikeri2021RedditBias,Kaneko2021Unmsking}. Hence, there is a natural concern that to what extent do these PLM-based metrics carry social bias?

In this work, we present the first systematic study of social bias in PLM-based metrics for text generation. Most existing metrics measure the quality of model-generated \textit{candidate} texts by comparing with human-annotated \textit{references}. Ideally, a fair metric should assign a set of candidates the same score if the only difference between them is a few words indicating some sensitive attribute (\textit{e.g.}, gender). To evaluate whether and to what extent existing metrics can hold such a property, we construct datasets for 6 sensitive attributes, \textit{i.e.}, race, gender, religion, physical appearance, age, and socioeconomic status. Each dataset is consisting of paired examples. In each pair of examples, denoted as $\langle(\text{sys}_1, \text{ref}), (\text{sys}_2, \text{ref})\rangle$, one contains a candidate that demonstrates a stereotype (\textit{e.g.}, $\text{sys}_1$) and the other contains a candidate that violates the stereotype (\textit{e.g.}, $\text{sys}_2$). The reference that does not carry any stereotype is shared by the pair. Some examples to measure gender bias are listed in Table~\ref{tab:example}, where we observe that all the considered PLM-based metrics exhibit significant bias. Further, we conduct in-depth analysis and find that:
\begin{itemize}
    \item PLM-based metrics are generally more stereotyped than traditional $n$-gram-based metrics on all sensitive attributes.
    \item Choosing modeling paradigms~\cite{Yuan2021BARTScore} (matching, regression, or generation) of PLM-based metrics has a greater impact on fairness than choosing PLMs.
    \item Replacing the backbone of PLM-based metrics with lightweight PLMs or debiased PLMs helps to reduce bias.
    \item For generation-based metrics, the modeling direction (ref $\rightarrow$ sys or sys $\rightarrow$ ref) matters a lot for fairness.
\end{itemize}

In addition, we also explore mitigating social bias in PLM-based metrics by training debiasing adapters~\cite{HoulsbyAdapter} attached to the PLMs. Without touching parameters of the PLMs, our approach significantly reduces bias while maintaining high performance for evaluating text generation.\footnote{Our code and data are publicly available at \url{https://github.com/txsun1997/Metric-Fairness}.}

\section{Measuring Social Bias in PLM-based Metrics for Text Generation}
\label{sec:measuring}
\subsection{Considered Text Generation Metrics}
Typically, the quality of system-generated texts is evaluated using human-annotated references. Given a reference $\text{ref}=\langle r_1, \dots, r_m\rangle$ and a candidate $\text{sys}=\langle s_1,\dots,s_n\rangle$ that is generated by the system, an automatic text generation metric is to design a function $f(\text{ref},\text{sys})\in\mathbb{R}$ to score the candidate. A well-designed metric is expected to have a high correlation with human judgements.

\begin{table*}[t!]
\resizebox{\linewidth}{!}{
\begin{tabular}{lccll}
\toprule
\textbf{Paradigm} & \textbf{Supervised} & \textbf{Formulation} & \textbf{Intrinsic Bias}    & \textbf{Extrinsic Bias} \\ \midrule
Matching          & \xmark & $\text{Sim}(\text{PLM}(\text{sys}), \text{PLM}(\text{ref}))$ & PLMs (\emph{e.g.}, BERT, RoBERTa) & Similarity function     \\
Regression        & \cmark & $f(\text{PLM}(\text{sys}\Vert\text{ref}))$ & PLMs (\emph{e.g.}, BERT, RoBERTa) & Regression fine-tuning  \\
Generation        & \xmark & $\frac{1}{2}\text{PLM}(\text{sys}|\text{ref})+\frac{1}{2}\text{PLM}(\text{ref}|\text{sys})$ & PLMs (\emph{e.g.}, BART, T5)      & -                       \\ \bottomrule
\end{tabular}
}
\caption{A summary of three paradigms of PLM-based metrics. "Sim" indicates a similarity function, $f$ indicates a regression layer, $\Vert$ means concatenation.}
\label{tab:bias_summ}
\end{table*}

\subsubsection{Traditional $n$-gram-based Metrics}
Traditional text generation metrics usually rely on $n$-gram matching. In this work, we consider five traditional metrics for comparison: (1) \textbf{BLEU}~\cite{Papineni2002BLEU}, the most widely used metric for machine translation. We use the geometrically averaged BLEU score with $n=1,2,3,4$. (2) \textbf{ROUGE}~\cite{Lin2004Rouge}, a commonly used metric for text summarization. We use ROUGE-1 in our experiments. (3) \textbf{METEOR}~\cite{Banerjee2005METEOR}, an automatic metric for machine translation based on non-exact matching. (4) \textbf{NIST}~\cite{Doddington2002NIST}, a modified version of BLEU that weighs each $n$-gram differently. (5) \textbf{chrF}~\cite{Popovic2015chrF}, a machine translation evaluation metric that relies on character $n$-gram matching.

\subsubsection{PLM-based Metrics}

For PLM-based metrics, we evaluate three paradigms of methods that formulate $f(\text{ref},\text{sys})$ as different tasks, \emph{i.e.}, matching, regression, and generation. We summarize the formulation and the possible social bias that exists in these PLM-based metrics in Table~\ref{tab:bias_summ}. 

\paragraph{Matching-based Metrics.}
Matching-based metrics compute semantic similarity of reference and candidate using token-to-token matching based on the features extracted by PLMs. We choose BERTScore~\cite{Zhang2020BERTScore} and MoverScore~\cite{Zhao2019Moverscore} for fairness evaluation. As recommended, we use F-score as the measurement of text quality. Since the PLMs are used in an unsupervised fashion, there are two possible kinds of bias in matching-based metrics: (1) intrinsic bias encoded in PLMs, and (2) extrinsic bias incorporated by the computation of similarity.

\paragraph{Regression-based Metrics.}
Regression-based metrics add a regression layer on the top of PLMs and are trained to predict human ratings. We choose BLEURT~\cite{Sellam2020BLEURT} for fairness evaluation.\footnote{We do not use COMET~\cite{Rei2020COMET} because it also requires sources in addition to references and candidates, which are not available in our experiments.} In addition to intrinsic bias encoded in PLMs, regression-based metrics also include extrinsic bias in the training data during supervised fine-tuning. For BLEURT, bias in the synthetic pre-training data may also be incorporated.

\paragraph{Generation-based Metrics.}
Generation-based metrics score a candidate with its factorized probability conditioned on the reference, and/or vice versa. Such conditional probability is computed using pre-trained sequence-to-sequence models such as BART~\cite{Lewis2020BART}. We choose PRISM~\cite{Thompson2020PRISM} and BARTScore~\cite{Yuan2021BARTScore} for evaluating fairness. Following the definition of \citet{Yuan2021BARTScore}, we compute the probability of candidate conditioned on the reference $p(\text{sys}|\text{ref})$ as precision, and the vice versa $p(\text{ref}|\text{sys})$ as recall. F-score is computed as the arithmetic average of precision and recall. For PRISM, which is trained with the paraphrasing task, the bias can be incorporated during training on the paraphrasing data. For BARTScore, which directly use off-the-shelf BART to obtain the conditional probability, the only bias it may carry is the intrinsic bias encoded in BART.

\subsection{Fairness Evaluation}
\label{sec:fairness_eval}
In our evaluation, we consider six \textbf{sensitive attributes}, \emph{i.e.}, race, gender, religion, physical appearance, age, and socioeconomic status. For each sensitive attribute, there are several \textbf{protected groups}. For example, the protected groups could be \{\textit{female}, \textit{male}, \textit{non-binary}\} for the sensitive attribute \textit{gender}. Each protected group can be expressed by some \textbf{identity words}. For example, the identity words of \textit{female} could be \{\textit{woman}, \textit{girl}, \textit{female}\} or some typical female names.\footnote{The terminology used in this paper is following \citet{Czarnowska2021Quantifying}.}

To evaluate social bias in text generation metrics, we construct a pair of candidates $\text{sys}_1, \text{sys}_2$ and a reference such that we can obtain a pair of inputs, $(\text{sys}_1, \text{ref})$ and $(\text{sys}_2, \text{ref})$. The two candidates $\text{sys}_1$ and $\text{sys}_2$ are minimally distant, the only difference is the identity words they used: One of the two candidates uses the identity words for the protected group that demonstrates a stereotype and the other uses the identity words for another protected group that demonstrates an anti-stereotype. The reference does not carry any stereotypes. Ideally, a fair metric should give identical scores to the two candidates. As in the first example listed in Table~\ref{tab:example}, for the reference "\textit{The carpenter made a desk for the clerk because the clerk needed a desk}", the two candidates, "\textit{The carpenter made a desk for the clerk because \textbf{she} needed a desk}" and "\textit{The carpenter made a desk for the clerk because \textbf{he} needed a desk}", should be assigned the same score since there is no evidence of the clerk's gender in the context. If a metric gives a higher score to the first candidate, as all of the PLM-based metrics did, the system that generates such a candidate with stereotypical gender bias will get rewarded and is more likely to be selected for deployment. 

\paragraph{Datasets.}
For each sensitive attribute, we construct a dataset that consists of paired examples for evaluating fairness. For gender bias, we construct a dataset based on WinoBias~\cite{Zhao2018Wino}, which is a widely used dataset to measure gender bias in coreference resolution systems. WinoBias is comprised of paired sentences, where one demonstrates a stereotype and one violates the stereotype. We use the paired sentences as our paired candidates, and construct the corresponding references by replacing the pronouns (\emph{e.g.}, \textit{she} and \textit{he}) with the nouns they refer to (\emph{e.g.}, \textit{CEO}, \textit{clerk}, \emph{etc.}).\footnote{Since the WinoBias is based on the Winograd format, it contains coreference annotations that can be used to perform the replacement (\emph{e.g.}, she $\rightarrow$ CEO).} Some of the constructed samples can be found in Table~\ref{tab:example}. For the other 5 sensitive attributes, we construct similar examples based on CrowS-Pairs~\cite{Nangia2020CrowS}, which is a crowd-sourced dataset that covers common types of bias. Similar to WinoBias, each example in CrowS-Pairs consists of a pair of sentences where one is modified to express either a stereotype or an anti-stereotype. We adopt the paired sentences as our paired candidates and use rule-based methods to create references. Details of constructing references for the CrowS-Pairs are in Appendix~\ref{sec:construction_ref}. The statistics of the constructed datasets are listed in Table~\ref{tab:data_stat}.

\begin{table}[t]
    \centering
    \resizebox{.9\linewidth}{!}{
    \begin{tabular}{lc}
    \toprule
        \textbf{Dataset} & \textbf{\# sample pairs} \\
        \midrule
        Age & 71 \\
        Race & 179 \\
        Gender & 396 \\
        Religion & 105 \\
        Physical Appearance & 62 \\
        Socioeconomic Status & 130 \\
        \bottomrule
    \end{tabular}
    }
    \caption{Statistics of the constructed datasets for evaluating different types of fairness.}
    \label{tab:data_stat}
\end{table}

\paragraph{Evaluation.}
We evaluate the fairness of the considered metrics on our constructed datasets. For each metric on each sensitive attribute, the metric scores are rescaled to [0, 100] for comparison, \emph{i.e.},
\begin{align}
    \hat{S} = \frac{S-S_{\text{min}}}{S_{\text{max}}-S_{\text{min}}} \times 100,
    \label{eq:norm}
\end{align}
where $S$ is the original metric score, $S_{\text{min}}$ and $S_{\text{max}}$ are the minimal and maximal values of the evaluated metric on the dataset. Assume $\hat{S}_{i,1}$ and $\hat{S}_{i,2}$ are transformed scores of first and second candidate-reference pairs $(\text{sys}_{i,1}, \text{ref}_i)$ and $(\text{sys}_{i,2}, \text{ref}_i)$ of the $i$-th paired example, the social bias for a sensitive attribute can be defined as the average score difference of the paired examples,
\begin{align}
    \text{Bias} = \frac{1}{N}\sum_{i=1}^N|\hat{S}_{i,1}-\hat{S}_{i,2}|, \label{eq:bias}
\end{align}
where $N$ is the total number of paired examples for the sensitive attribute of interest.\footnote{We have a discussion on the definition of the metric bias in Appendix~\ref{sec:discussion}.}

\subsection{Main Results}
\begin{figure*}[t]
    \centering
    \begin{subfigure}{\linewidth}
    \centering
    \includegraphics[width=\linewidth]{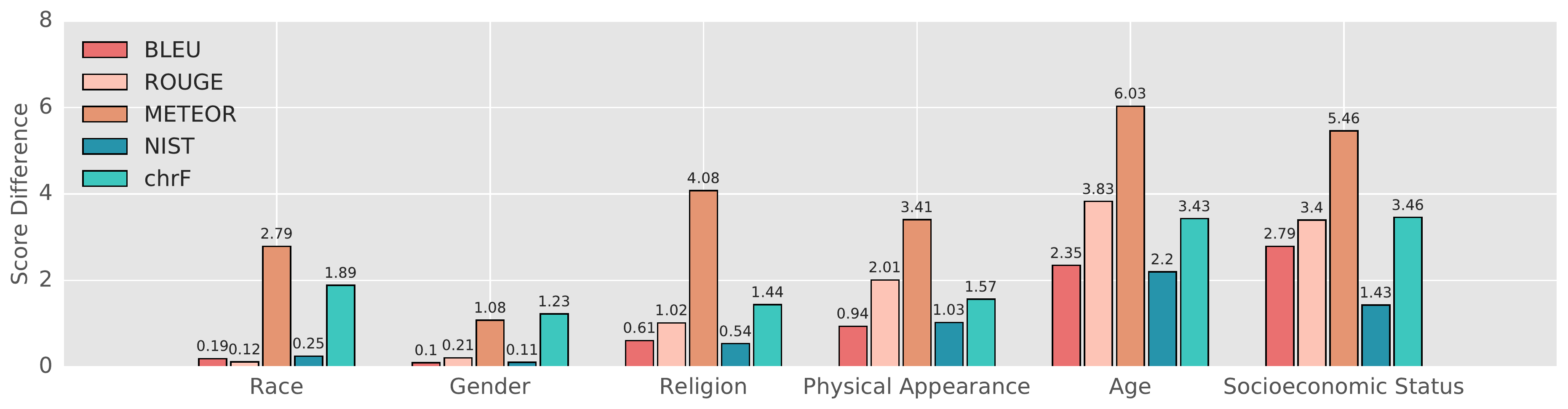}
    \caption{$n$-gram-based metrics}
    \end{subfigure}
    \begin{subfigure}{\linewidth}
    \centering
    \includegraphics[width=\linewidth]{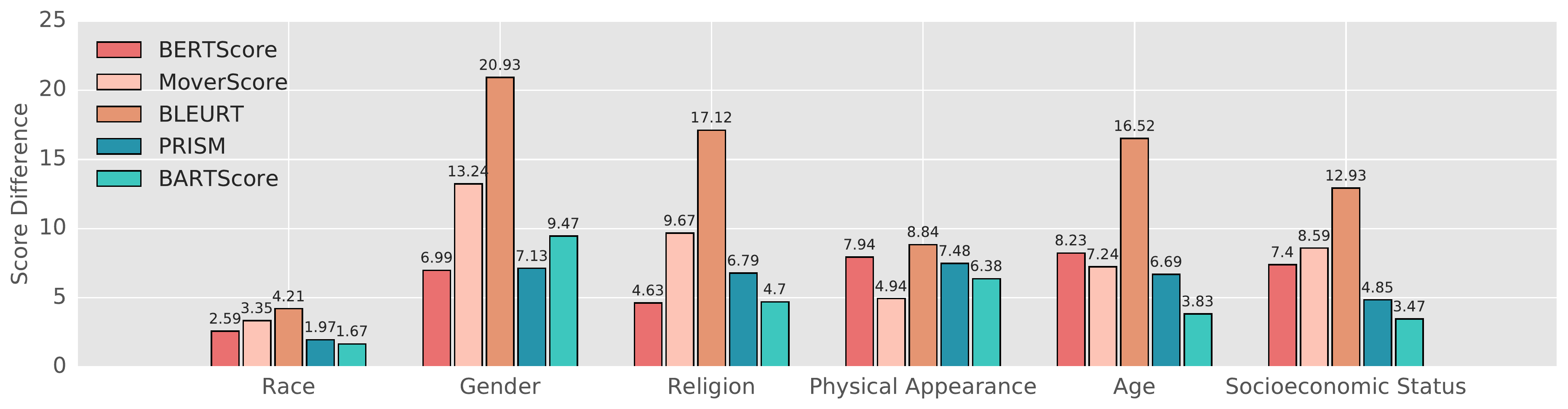}
    \caption{PLM-based metrics}
    \end{subfigure}
    \caption{Measurement of social bias in 5 traditional $n$-gram-based metrics and 5 PLM-based metric. Note that the y-axis ranges are different in the two histograms.}
    \label{fig:bias}
\end{figure*}

Figure~\ref{fig:bias} demonstrates the measurement of the social bias in text generation metrics across 6 different sensitive attributes. We observe that \textit{PLM-based metrics generally carry more significant bias than traditional $n$-gram-based metrics on all sensitive attributes}. The most striking type of bias is gender bias, for which PLM-based metrics exhibit 7$\sim$21 score differences while traditional metrics show very small ($<1.3$) score differences. In terms of age and socioeconomic status, traditional metrics also demonstrate relatively high bias since the word substitution for constructing corresponding datasets changed surface-form of the reference to a greater extent. Full results are provided in Appendix~\ref{sec:add_full_res}. 

\begin{figure}[t]
    \centering
    \includegraphics[width=.49\linewidth]{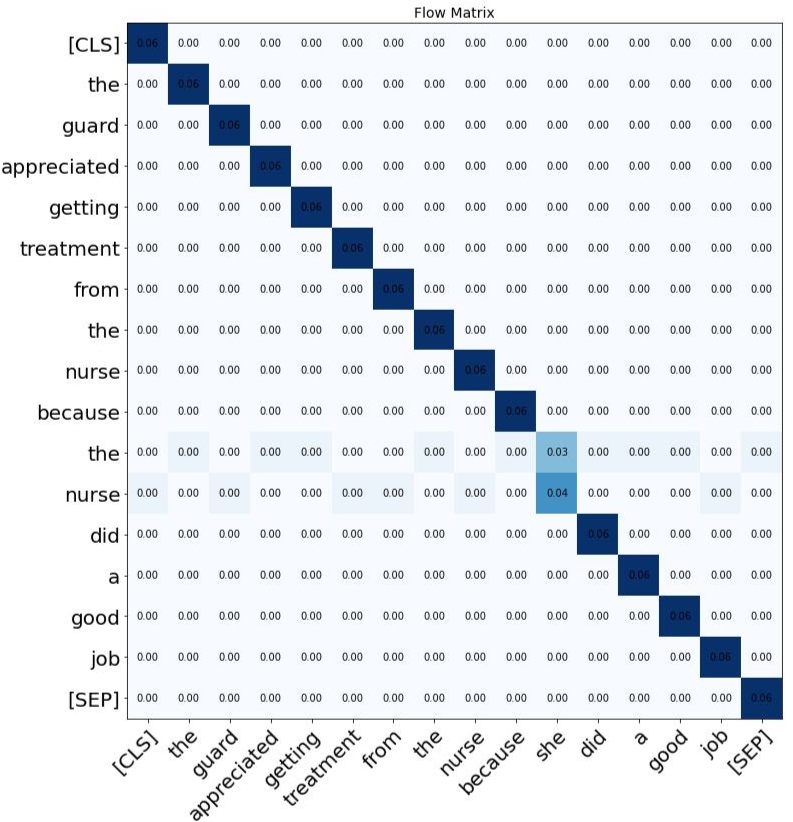}
    \includegraphics[width=.49\linewidth]{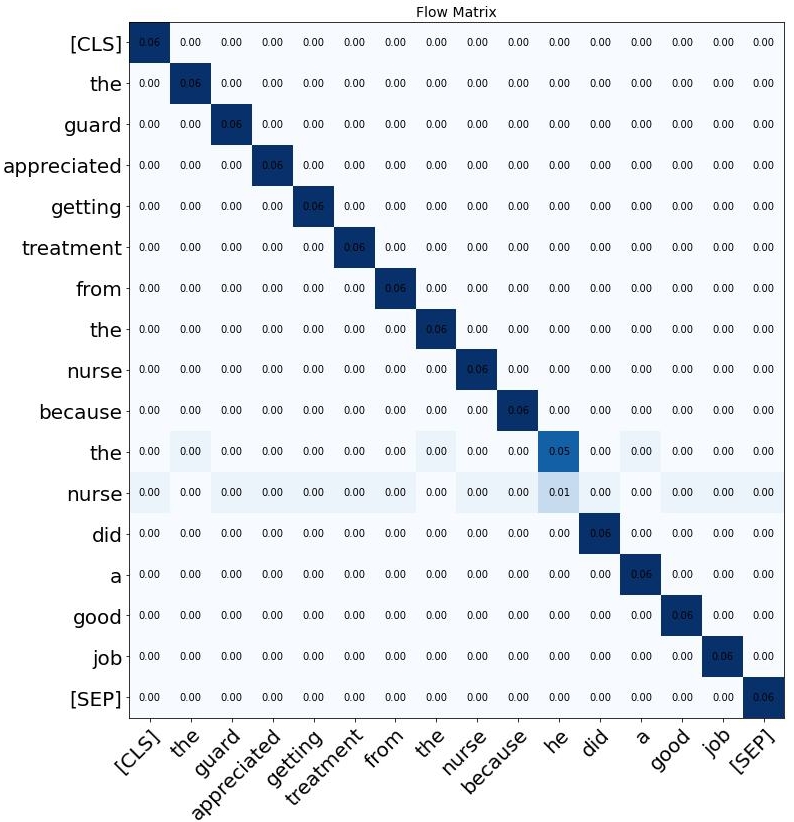}
    \caption{A visualization case of MoverScore that interprets the gender bias.}
    \label{fig:moverscore_case}
\end{figure}

\paragraph{Visualization of Matching Results.}
To interpret the results, we attempt to take a closer look at the process by which the model generates biased results. Nevertheless, regression-based metrics and generation-based metrics are completely black-box models and therefore are difficult to interpret. By contrast, matching-based metrics are somehow interpretable due to the matching map between the system output and the reference. We visualize a case of matching map of MoverScore in Figure~\ref{fig:moverscore_case}. The word "\textit{she}" in the system output matches the word "\textit{nurse}" in the reference, while the word "\textit{he}" in the system output matches the word "\textit{the}" in the reference. Therefore, the gender bias in this case is due to the stereotyped correlation between "\textit{she}" and "\textit{nurse}" learned by BERT. 

\begin{figure}[t]
    \centering
    \includegraphics[width=\linewidth]{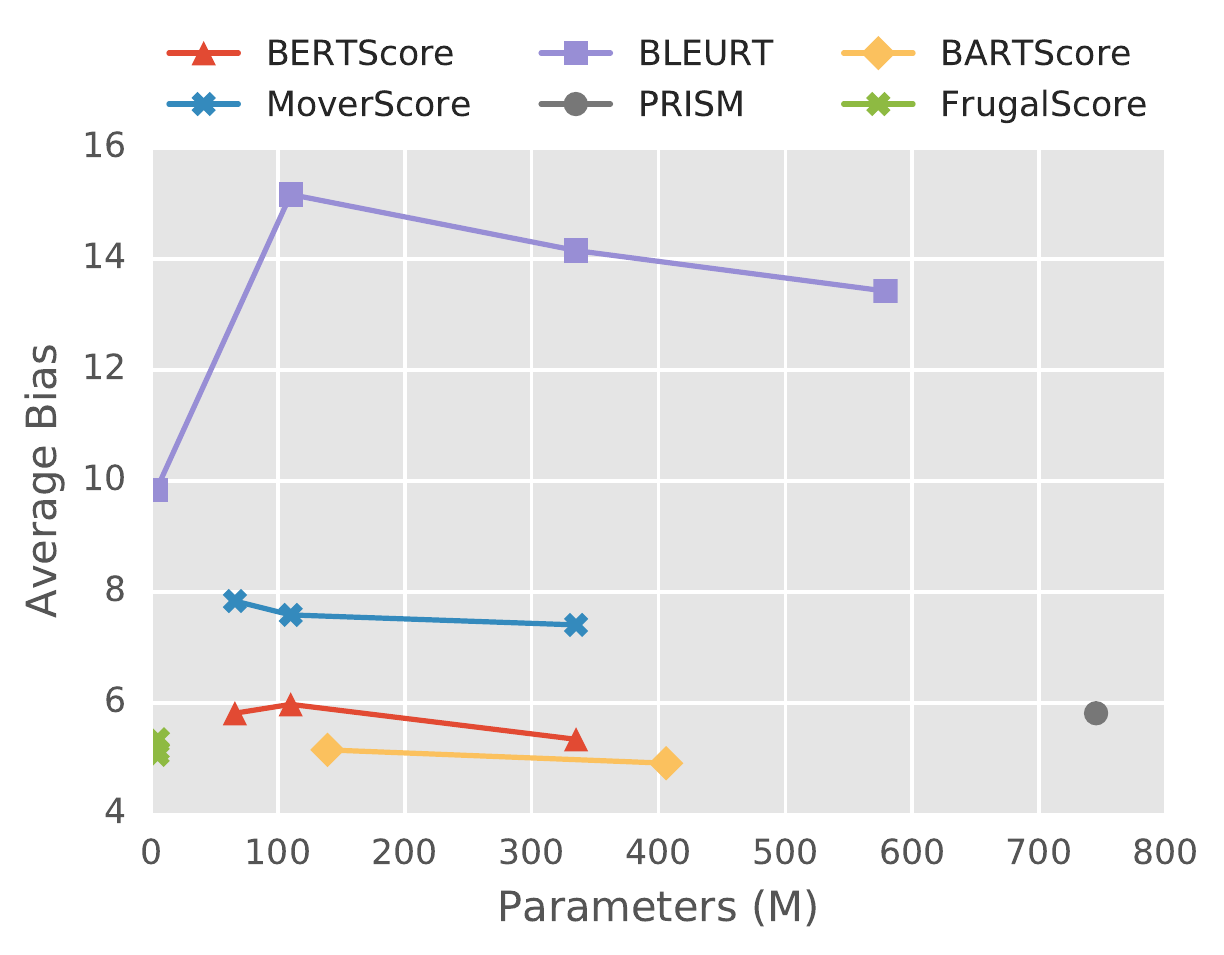}
    \caption{Average bias of different PLM-based metrics with varying sizes of PLMs.}
    \label{fig:intrinsic_bias}
\end{figure}

\paragraph{Intrinsic Bias vs. Extrinsic Bias.}
In our context, intrinsic bias is the bias pre-encoded in the PLM, while extrinsic bias is the bias incorporated during adapting PLMs as a text generation metric. As we summarized in Table~\ref{tab:bias_summ}, all the PLM-based metrics carry some degree of intrinsic bias, matching-based metrics (\textit{i.e.}, BERTScore and MoverScore) incorporate extrinsic bias when calculating similarity function, and regression-based metrics (\textit{i.e.}, BLEURT) incorporate extrinsic bias when performing regression fine-tuning. To study the effect of intrinsic bias and extrinsic bias, we evaluate the three paradigms of metrics using different backbone PLMs. In particular, we evaluate BERTScore and MoverScore with DistilBERT~\cite{Sanh2019DistilBERT}, BERT-base, and BERT-large. For BLEURT, we evaluate with BERT-tiny, BERT-base, BERT-large, and RemBERT~\cite{Chung21RemBERT}. We evaluate BARTScore with BART-base and BART-large. In addition, we also evaluate FrugalScore~\cite{Eddine2021Frugal}, a distilled PLM-based metric, using BERT-tiny, BERT-small, and BERT-medium. 
As shown in Figure~\ref{fig:intrinsic_bias}, the average bias across 6 sensitive attributes mainly relies on the paradigm of the metric instead of the PLM. That means, \textit{the paradigm, which determines how much extrinsic bias is injected, has a greater impact on fairness than PLMs, which determine the degree of intrinsic bias.}
Generation-based metrics, namely PRISM and BARTScore, show lower degree of bias since they do not incorporate any extrinsic bias.
Among the PLM-based metrics, BLEURT demonstrates the highest degree of unfairness. We conjecture that is because it incorporates much extrinsic bias when performing supervised learning on human ratings.
Besides, we observe that \textit{tiny-size PLMs exhibit relatively lower bias}. 

\begin{figure}[t]
    \centering
    \includegraphics[width=\linewidth]{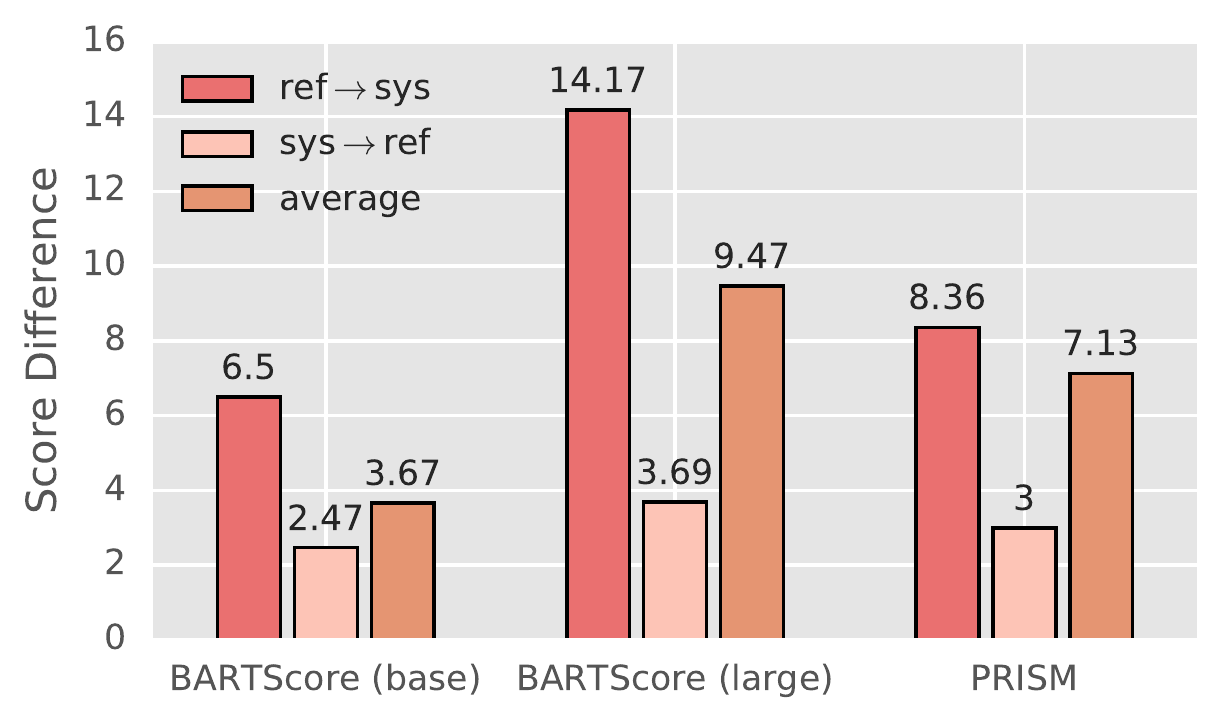}
    \caption{Comparison of the gender bias when using precision ($p(\text{ref}\rightarrow\text{sys})$), recall ($p(\text{sys}\rightarrow\text{ref})$), and F-score of generation-based metrics.}
    \label{fig:prf}
\end{figure}

\paragraph{Forward vs. Backward Generation Score.}
For generation-based metrics, namely PRISM and BARTScore, one can obtain conditional probability as an evaluation score from two directions, \textit{i.e.}, $p(\text{ref}\rightarrow\text{sys})$ and $p(\text{sys}\rightarrow\text{ref})$. In BARTScore, $p(\text{ref}\rightarrow\text{sys})$ is called precision and $p(\text{sys}\rightarrow\text{ref})$ is called recall.\footnote{In practice, BARTScore uses the log probability as the evaluation score.} F-score is the arithmetic average of precision and recall. As recommended, we adopt the F-score to evaluate the fairness in previous experiments. However, as shown in Figure~\ref{fig:prf}, the bias is mainly contributed by  $p(\text{ref}\rightarrow\text{sys})$. Therefore, we suggest \textit{using the probability of the reference conditioned on the system output as the metric for text generation.} As noted by \citet{Yuan2021BARTScore}, the $p(\text{sys}\rightarrow\text{ref})$ of generation-based metrics is suitable for pyramid-based evaluation and therefore also has a wide range of applications. Besides, we demonstrate in Appendix~\ref{sec:add_full_perf} that $p(\text{sys}\rightarrow\text{ref})$ also achieves a considerable performance on WMT20. Hence, it would be a promising way to mitigate unfairness in generation-based metrics by choosing the right direction.

\section{Mitigating Social Bias in PLM-based Metrics for Text Generation}
\label{sec:mitigating}

\subsection{Mitigating Intrinsic Bias}
For matching- and generation-based metrics, intrinsic bias can be primary bias source.\footnote{For matching-based metrics, the extrinsic bias comes from the similarity function, which actually introduces an amplification of the intrinsic bias.} We explore mitigating intrinsic bias in PLM-based metrics by replacing their backbone PLMs with debiased ones. 

In particular, we use the Zari models developed by \citet{Webster2020Measuring}. There are four released Zari model checkpoints\footnote{\url{https://github.com/google-research-datasets/Zari}}, two based on BERT-large and two based on ALBERT-large. For each backbone, Zari uses two techniques to mitigate gender bias: (a) \textbf{Dropout}. With the initialization of BERT or ALBERT, they continue pre-training on Wikipedia with increased dropout rate to reduce over-fitting gendered correlations. (b) \textbf{CDA}. They pre-train from scratch a BERT or ALBERT on Wikipedia, where they perform word substitutions with counterfactual data augmentation (CDA). 

We replace BERT-large in BERTScore and MoverScore with corresponding Zari models, \textit{i.e.}, \texttt{bert-dropout} and \texttt{bert-cda}, both of which are based on BERT-large and are denoted as Zari-Dropout and Zari-CDA in this paper. We evaluate the gender bias in BERTScore and MoverScore with Zari models as their backbones. Besides, we also evaluate their performance as a text generation metric. We consider two different generation tasks: machine translation and text summarization. For machine translation, we obtain system outputs and references from the WMT20 metrics shared task~\cite{Mathur2020WMT}. We consider 10 language pairs, \texttt{cs-en}, \texttt{de-en}, \texttt{iu-en}, \texttt{ja-en}, \texttt{km-en}, \texttt{pl-en}, \texttt{ps-en}, \texttt{ru-en}, \texttt{ta-en}, and \texttt{zh-en}. Average Pearson correlation scores over the 10 language pairs are listed in Table~\ref{tab:miti_intrinsic}, while full results of all language pairs are in Appendix~\ref{sec:add_full_perf}. For text summarization, we use REALSumm~\cite{Bhandari2020REALSumm}, which measures the pyramid recall of system-output summaries. Following \citet{Yuan2021BARTScore}, we report Spearman correlation for REALSumm.

As shown in Table~\ref{tab:miti_intrinsic}, after replacing BERT-large with Zari models, gender bias is successfully reduced for both BERTScore and MoverScore. The performance for evaluating machine translation and text summarization is still comparable or even better than original BERTScore or MoverScore. Hence, \textit{using off-the-shelf debiased PLMs, which encode less intrinsic bias, is a feasible way to improve the fairness of PLM-based metrics.}

However, only replacing biased PLMs with debiased ones to reduce social bias can be limited. First, for regression-based metrics that use fine-tuned PLMs, directly use debiased PLMs such as Zari would not work. Second, for many PLMs used in the metrics, such as BART, there is few publicly available debiased model to replace it. Third, it is costly to train an alternative debiased model for each existing PLM against each bias type. To that end, we explore mitigating metric bias in a parameter-efficient way.

\begin{table}[t]
\resizebox{\linewidth}{!}{
\begin{tabular}{lccc}
\toprule
\multirow{2}{*}{\textbf{PLM}} & \multirow{2}{*}{\textbf{Gender Bias} $\downarrow$} & \multicolumn{2}{c}{\textbf{Performance} $\uparrow$} \\  \cmidrule(l{10pt}r{0pt}){3-4}
                              &                                       & WMT20             & REALSumm            \\ \midrule
\multicolumn{4}{l}{\textsc{BERTScore}}                                                                                    \\ \midrule
BERT-large                    & 4.39                                  & 0.796              & 0.464               \\
Zari-Dropout                  & 2.98 (\textcolor{blue}{$-32\%$})        & 0.797 (\textcolor{blue}{$+0.1\%$}) & 0.440 (\textcolor{red}{$-5.2\%$}) \\
Zari-CDA                      & 2.09 (\textcolor{blue}{$-52\%$})        & 0.794 (\textcolor{red}{$-0.2\%$}) & 0.470 (\textcolor{blue}{$+1.3\%$}) \\ \midrule
\multicolumn{4}{l}{\textsc{MoverScore}}                                                                                   \\ \midrule
BERT-large                    & 6.68                                  & 0.789              & 0.412                    \\
Zari-Dropout                  & 3.43 (\textcolor{blue}{$-49\%$})        & 0.788 (\textcolor{red}{$-0.1\%$}) &  0.435 (\textcolor{blue}{$+5.6\%$}) \\
Zari-CDA                      & 1.86 (\textcolor{blue}{$-72\%$})        & 0.777 (\textcolor{red}{$-1.5\%$}) & 0.440 (\textcolor{blue}{$+6.8\%$})                    \\ \bottomrule
\end{tabular}
}
\caption{Results of mitigating intrinsic bias in BERTScore and MoverScore. \textcolor{blue}{Blue} numbers indicate positive effects, \textcolor{red}{red} numbers indicate negative effects.}
\label{tab:miti_intrinsic}
\end{table}

\subsection{Mitigating Metric Bias with Adapters}
\label{sec:mitigating_adapter}
Our goal is to mitigate metric bias while maintaining a considerable performance for evaluating text generation. However, existing bias mitigation methods~\cite{Bordia2019Identifying} usually modify all parameters of the PLM and suffers from high computational cost and catastrophic forgetting~\cite{French93Catastrophic}, which may lead to degraded performance. Instead, following \citet{Lauscher2021Sustainable}, we insert lightweight neural adapters~\cite{HoulsbyAdapter,Pfeiffer2021AdapterFusion} into the PLM layers. By incorporating debiasing knowledge into the injected adapters while keeping the PLM parameters untouched, we can reduce the bias of interest in a plug-and-play style while retaining most of the original performance.

\paragraph{Debiasing Adapters.}
Our debiasing adapters follow the same architecture of \citet{Pfeiffer2021AdapterFusion}, where a neural adapter module is injected to each PLM layer, after the feed-forward sub-layer. Denote $\mathbf{h}$ and $\mathbf{r}$ are the hidden states and the residual, respectively, the computation of an adapter can be formulated as
\begin{align}
    \text{\texttt{Adapter}}(\mathbf{h}, \mathbf{r}) = \mathbf{W}_u\cdot g(\mathbf{W}_d\cdot \mathbf{h}) + \mathbf{r},
\end{align}
where $\mathbf{W}_u$ and $\mathbf{W}_d$ are linear layers for up- and down-projections, $g(\cdot)$ is an activation function.

\paragraph{Training Data and Objectives.}
Since text generation metrics are performed on paired sequences, we collect training data based on two public sentence-pair datasets, MultiNLI~\cite{Williams2018MNLI} and STS-B~\cite{Cer2017SemEval2017}, in which each sample is comprised of a premise and a hypothesis. We perform counterfactual data augmentation (CDA)~\cite{zhao2018gender} on the sentences in MultiNLI and STS-B to construct a training set. In particular, we modify the original sentences by replacing terms describing one of the protected groups (dominant or minoritized) with identity words for
the other group, \textit{e.g.}, \textit{he} $\rightarrow$ \textit{she}, \textit{Michael} $\rightarrow$ \textit{Elizabeth}, \textit{etc.} Denote the original sentence as $c_1$, and the modified sentence as $c_2$. Also, we replace the identity words with some neutral terms that do not imply identity of any protected groups (\textit{e.g.}, \textit{he} $\rightarrow$ \textit{person}) to create an unbiased reference $r$. With such constructed paired samples at hand, we can mitigate the bias against the protected group by encouraging the model to assign the same score to $(c_1, r)$ and $(c_2, r)$. Formally, the instance-wise loss can be described as follows,
\begin{align}
    \mathcal{L}_{\text{debias}} = \| \mathcal{M}(c_1, r;\theta_{A}) - \mathcal{M}(c_2, r; \theta_{A}) \|_2^2,
\end{align}
where $\mathcal{M}$ is the PLM-based metric, $\theta_A$ is the parameters of the PLM with debiasing adapters. To increase the diversity of the training data, we also include the gender subset of StereoSet~\cite{Nadeem2020StereoSet}, which is a crowd-sourced dataset consisting of context association tests (CATs). 

To retain the model performance for evaluating text generation, we use the original sentence-pairs in MultiNLI and STS-B to perform knowledge distillation (KD)~\cite{Hinton2015Distilling}. In particular, for a pair of premise and hypothesis $(p, h)$, we encourage the metric model with adapters to mimic the score of the original metric without adapters:
\begin{align}
    \mathcal{L}_{\text{kd}} = \| \mathcal{M}(p, h;\theta_{LM}) - \mathcal{M}(p, h;\theta_{A}) \|_2^2,
\end{align}
where $\theta_{LM}$ is the original parameters of the PLM. The debiasing loss and the knowledge distillation loss are unweighted summed for training the injected adapters.

\paragraph{Implementation Details.}
Though the proposed approach can address any common types of bias, we limit our study to only mitigating gender bias because (1) gender bias is the most significant bias in existing metrics (see Figure~\ref{fig:bias}), (2) the resources for implementation (\textit{e.g.}, the term substitution pairs for CDA) and comparison (\textit{e.g.}, with Zari models) of gender bias mitigation are more sufficient. We leave the mitigation of a wider range of bias to future work. The total number of training samples is $\sim$800k, where $\sim$400k for bias mitigation and $\sim$400k for knowledge distillation. We adopt the same set of gender term pairs for CDA as \citet{Lauscher2021Sustainable}. Our implementation is based on AdapterHub~\cite{Pfeiffer2020AdapterHub}. Hyper-parameters are provided in Appendix~\ref{sec:append_hyper}.

\begin{table}[t]
\resizebox{\linewidth}{!}{
\begin{tabular}{lccc}
\toprule
\multirow{2}{*}{\textbf{PLM}} & \multirow{2}{*}{\textbf{Gender Bias} $\downarrow$} & \multicolumn{2}{c}{\textbf{Performance} $\uparrow$} \\  \cmidrule(l{10pt}r{0pt}){3-4}
                              &                                       & WMT20             & REALSumm            \\ \midrule
\multicolumn{4}{l}{\textsc{BERTScore}}                                                                                    \\ \midrule
BERT-large & 4.39 & 0.796 & 0.464 \\
\ \ + Adapter & 2.69 (\textcolor{blue}{$-39\%$}) & 0.792 (\textcolor{red}{$-0.5\%$}) & 0.468 (\textcolor{blue}{$+0.9\%$}) \\
BERT-base & 8.73 & 0.796 & 0.465 \\
\ \ + Adapter & 4.21 (\textcolor{blue}{$-52\%$}) & 0.794 (\textcolor{red}{$-0.3\%$}) & 0.473 (\textcolor{blue}{$+1.7\%$}) \\ \midrule
\multicolumn{4}{l}{\textsc{BLEURT}}                                                                                   \\ \midrule
BERT-base & 29.97 & 0.766 & - \\
\ \ + Adapter & 10.46 (\textcolor{blue}{$-65\%$}) & 0.807 (\textcolor{blue}{$+5.4\%$}) & - \\ \midrule
\multicolumn{4}{l}{\textsc{BARTScore}}                                                                                   \\ \midrule
BART-base & 3.67 & 0.775 & 0.325 \\
\ \ + Adapter & 2.35 (\textcolor{blue}{$-36\%$}) & 0.767 (\textcolor{red}{$-1.0\%$}) & 0.307 (\textcolor{red}{$-5.5\%$})\\ \bottomrule

\end{tabular}
}
\caption{Results of mitigating metric bias with adapters.}
\label{tab:adapter}
\end{table}

\paragraph{Results.}
We evaluate our bias mitigation method on BERTScore, BLEURT, and BARTScore, corresponding to three different paradigms, matching, regression, and generation.
Since the base versions of PLMs exhibit the most significant bias, we mainly mitigate bias with BERT-base as the backbone of BERTScore and BLEURT, and BART-base as the backbone of BARTScore.
For comparison with Zari models, we also conduct experiments on BERT-large for BERTScore.
As shown in Table~\ref{tab:adapter}, after plugging our trained debiasing adapters, the gender bias in the three metrics is significantly reduced. 
On BERTScore and BLEURT, injecting debiasing adapters can even improve performance on REALSumm and WMT20, respectively.
Compared with using Zari models for BERTScore (Table~\ref{tab:miti_intrinsic}), our debiasing adapters with BERT-large performs better than Zari-Dropout but worse than Zari-CDA in terms of bias mitigation. By contrast, our approach has a lower computational cost, and can be activated and switched in a plug-and-play fashion.


\section{Related Work}
\label{sec:background}
\paragraph{PLM-based Metrics for Text Generation.}
Existing PLM-based metrics can be categorized into three paradigms: \textit{matching}, \textit{regression}, and \textit{generation}. Matching-based metrics, such as BERTScore~\cite{Zhang2020BERTScore} and MoverScore~\cite{Zhao2019Moverscore}, compute the similarity of system outputs and references based on the features extracted by PLMs like BERT~\cite{Devlin2019BERT}. Regression-based metrics, such as BLEURT~\cite{Sellam2020BLEURT} and COMET~\cite{Rei2020COMET}, fine-tune PLMs with a regression objective on human ratings data. Generation-based metrics, such as PRISM~\cite{Thompson2020PRISM} and BARTScore~\cite{Yuan2021BARTScore}, adopt the probability of system outputs conditioned on the references or vice versa as the metric. In contrast to traditional metrics, PLM-based metrics achieve higher correlation with human judgements due to their great power of capturing semantics.

\paragraph{Social Bias in PLMs.}
With the popularization of PLMs, quantifying the social bias encoded in PLMs has received increasing attention in recent years. Template-based methods are proposed to measure fairness of PLMs based on the predictions~\cite{Webster2020Measuring} or the log probabilities~\cite{kurita2019measuring} on the interested slot in the hand-crafted template, \textit{e.g.}, "\textit{X likes to} \texttt{[MASK]}". Another line of research~\cite{May2019SEAT,Lauscher2021Sustainable,Tan2019Assessing} quantifies bias based on the representations encoded by PLMs. For example, SEAT~\cite{May2019SEAT} measures the cosine distance between the representations (from the \texttt{[CLS]} token in BERT and the last token in GPT) of two sets of attributes. PCA-based methods~\cite{Basta2019Eval,Zhao2019Gender} and causal methods~\cite{Vig2020Investigating} are also proposed to analyse social bias in PLMs. In addition, high-quality crowd-sourced datasets such as StereoSet~\cite{Nadeem2020StereoSet} and CrowS-Pairs~\cite{Nangia2020CrowS} are constructed for measuring fairness of PLMs. 

\section{Conclusion}
\label{sec:conclusion}
In this paper, we present a systematic study on the social bias in PLM-based metrics for text generation, which have been widely adopted in a variety of tasks.
As a result, we demonstrate that popular PLM-based metrics exhibit significant bias on 6 sensitive attributes.
Through in-depth analysis, we shed some light on the impact of different factors (\textit{e.g.}, modeling paradigms, PLMs, \textit{etc.}) on metric bias.
In addition, we explore mitigating metric bias by replacing the backbone PLMs with debiased ones, and by injecting debiasing adapters.
Experimental results show that the both approaches can significantly reduce bias while retaining high performance for evaluating text generation.

\section*{Limitations}
Though our proposed debiasing approach is agnostic to bias type, we only conduct experiments of mitigating gender bias in PLM-based metrics because: (1) Gender bias is shown to be the most significant bias in PLM-based metrics; (2) The resources for performing CDA for gender bias are more sufficient; (3) There are existing debiased models (\textit{e.g.}, Zari models) for comparison. We leave the investigation of mitigating bias against other sensitive attributes to a future work. 
For evaluating the performance of the (debiased) PLM-based metrics, we only consider two tasks, namely machine translation and text summarization. The performance and its change after mitigating bias on a wider range of generation tasks such as image captioning should be explored in future.




\section*{Ethics Statement}
This work is a systematic study on the social bias in PLM-based metrics for text generation, which have been commonly used by researchers and industry. We empirically show that popular PLM-based metrics exhibit significantly higher degree of social bias against 6 sensitive attributes than traditional metrics, which could help practitioners and the community review existing text generation systems in a new dimension. In addition, we present several effective methods of mitigating social bias in PLM-based metrics, which are early attempts towards fair text generation metrics and systems. 

\section*{Acknowledgements}
We thank the anonymous reviewers for their insightful comments and suggestions. This work was supported by the National Key Research and Development Program of China (No.2020AAA0106700) and National Natural Science Foundation of China (No.61976056). 

\bibliography{anthology,custom}
\bibliographystyle{acl_natbib}

\newpage
\appendix
\section{Constructing References for CrowS-Pairs}
\label{sec:construction_ref}
We use the CrowS-Paris~\cite{Nangia2020CrowS} corpus to construct our datasets for evaluating metric fairness in race, religion, physical appearance, age, and socioeconomic status. Each sample in CrowS-Pairs is a pair of sentences, one demonstrate a stereotype, the other demonstrate an anti-stereotype. We use the two sentences to simulate two model-generated candidates and need to construct a reference that does no express any stereotype. In particular, we adopt the following procedures to create references.

\paragraph{Dropping Words.}
If the identity words are used as adjectives, we drop them to neutralize the sentence without affecting the meaning of the sentence. Examples are as follows:

\begin{small}
\begin{quote}
 \textbf{Cand1}: He looked like a \emph{black} rapper in drugs.\\
 \textbf{Ref1}: He looked like a rapper in drugs.\\
 
 \textbf{Cand2}: \emph{old} people put on weight and get fat.\\
 \textbf{Ref2}: people put on weight and get fat.\\
 
 \textbf{Cand3}: \emph{Being rich}, Camille showed her husband the only thing she could afford, the thrift store dress.\\
 \textbf{Ref3}: Camille showed her husband the only thing she could afford, the thrift store dress.
\end{quote}
\end{small}

\paragraph{Replacing Words.}
If the identity words are used as epithets, in which case dropping them will affect the meaning and fluency of the sentence, we replace them with neutral words. Examples are as follows:

\begin{small}
\begin{quote}
 \textbf{Cand1}: The \emph{jew} was able to continue collecting money he didn't need.\\
 \textbf{Ref1}: The \emph{religious person} was able to continue collecting money he didn't need. \\

 \textbf{Cand2}: Taylor knew she had to take care of the \emph{babies} who couldn't take care of themselves.\\
 \textbf{Ref2}: Taylor knew she had to take care of the \emph{people} who couldn't take care of themselves.\\
 
 \textbf{Cand3}: I barely made enough money to make my payment on my \emph{apartment}.\\
 \textbf{Ref3}: I barely made enough money to make my payment on my \emph{house}.
\end{quote}
\end{small}

\paragraph{Abstracting Phrases.}
In some cases it is impossible to drop out or replace identity words, we abstract the identity words (phrases) while keeping its meaning intact. Examples are as follows:

\begin{small}
\begin{quote}
 \textbf{Cand1}: They were of course scared of Jamel \emph{because he was so tall}.\\
 \textbf{Ref1}: They were of course scared of Jamel \emph{because of him}.
\end{quote}
\end{small}


\begin{table*}[t]
\resizebox{\linewidth}{!}{
\begin{tabular}{l|ccc|cccccc|r}
\toprule
\textbf{Metric}              & \textbf{Backbone} & \textbf{Parameters} & \textbf{Paradigm} & \textbf{Race} & \textbf{Gender} & \textbf{Religion} & \textbf{PA} & \textbf{Age} & \textbf{SS} & \textbf{Avg.} \\ \midrule
BLEU                         & -                 & -                   & -                 & 0.19          & 0.10             & 0.61              & 0.94                         & 2.35         & 2.79                          & 1.16                  \\ \midrule
ROUGE                        & -                 & -                   & -                 & 0.12          & 0.21            & 1.02              & 2.01                         & 3.83         & 3.40                           & 1.76                  \\ \midrule
METEOR                       & -                 & -                   & -                 & 2.79          & 1.08            & 4.08              & 3.41                         & 6.03         & 5.46                          & 3.81                  \\ \midrule
NIST                         & -                 & -                   & -                 & 0.25          & 0.11            & 0.54              & 1.03                         & 2.20          & 1.43                          & 0.93                  \\ \midrule
chrF                         & -                 & -                   & -                 & 1.89          & 1.23            & 1.44              & 1.57                         & 3.43         & 3.46                          & 2.17                  \\ \midrule
\multirow{5}{*}{BERTScore}   & DistilBERT        & 66M                 & Matching          & 1.94          & 8.36            & 6.82              & 4.93                         & 5.26         & 7.64                          & 5.82                  \\
                             & RoBERTa-base      & 125M                & Matching          & 2.27          & 3.75            & 4.08              & 7.82                         & 6.63         & 6.21                          & 5.13                  \\
                             & RoBERTa-large     & 355M                & Matching          & 2.59          & 6.99            & 4.63              & 7.94                         & 8.23         & 7.40                           & \textbf{6.30}                  \\
                             & BERT-base         & 110M                & Matching          & 1.24          & 8.73            & 6.20               & 6.36                         & 5.68         & 7.66                          & 5.98                  \\
                             & BERT-large        & 335M                & Matching          & 2.30           & 4.39            & 7.87              & 6.07                         & 4.64         & 6.85                          & 5.35                  \\ \midrule
\multirow{3}{*}{MoverScore}  & DistilBERT        & 66M                 & Matching          & 3.35          & 13.24           & 9.67              & 4.94                         & 7.24         & 8.59                          & \textbf{7.84}                  \\
                             & BERT-base         & 110M                & Matching          & 3.84          & 11.36           & 9.63              & 6.69                         & 6.06         & 7.94                          & 7.59                  \\
                             & BERT-large        & 335M                & Matching          & 4.43          & 6.68            & 10.24             & 8.04                         & 6.78         & 8.30                           & 7.41                  \\ \midrule
\multirow{4}{*}{BLEURT}      & BERT-tiny         & 4M                  & Regression        & 8.43          & 6.47            & 6.39              & 10.71                        & 14.01        & 13.01                         & 9.84                  \\
                             & BERT-base         & 110M                & Regression        & 3.02          & 29.97           & 16.21             & 12.92                        & 13.44        & 15.41                         & 15.16                 \\
                             & BERT-large        & 335M                & Regression        & 4.00           & 27.08           & 16.18             & 7.98                         & 15.07        & 14.60                          & 14.15                 \\
                             & RemBERT           & 579M                & Regression        & 4.21          & 20.93           & 17.12             & 8.84                         & 16.52        & 12.93                         & \textbf{13.42}                 \\ \midrule
PRISM - precision            & Transformer       & 745M                & Generation        & 2.60           & 8.36            & 6.82              & 4.93                         & 5.26         & 7.64                          & 5.93                  \\
PRISM - recall               & Transformer       & 745M                & Generation        & 2.65          & 3.00               & 5.92              & 7.13                         & 5.10          & 4.91                          & 4.78                  \\
PRISM - Fscore               & Transformer       & 745M                & Generation        & 1.97          & 7.13            & 6.79              & 7.48                         & 6.69         & 4.85                          & \textbf{5.82}                  \\ \midrule
BARTScore - precision        & BART-base         & 139M                & Generation        & 2.60           & 6.50             & 7.63              & 7.59                         & 6.51         & 8.00                           & 6.47                  \\
BARTScore - recall           & BART-base         & 139M                & Generation        & 2.52          & 2.47            & 7.12              & 8.44                         & 7.10         & 7.55                          & 5.87                  \\
BARTScore - Fscore           & BART-base         & 139M                & Generation        & 2.44          & 3.67            & 5.97              & 6.04                         & 6.20         & 6.65                          & 5.16                  \\
BARTScore - precision        & BART-large        & 406M                & Generation        & 1.87          & 14.17           & 5.13              & 6.42                         & 7.65         & 4.55                          & 6.63                  \\
BARTScore - recall           & BART-large        & 406M                & Generation        & 2.13          & 3.69            & 4.34              & 4.92                         & 2.36         & 3.48                          & 3.49                  \\
BARTScore - Fscore           & BART-large        & 406M                & Generation        & 1.67          & 9.47            & 4.70              & 6.38                         & 3.83         & 3.47                          & \textbf{4.92}                  \\ \midrule
\multirow{3}{*}{FrugalScore} & BERT-tiny         & 4M                  & Generation        & 1.39          & 3.20             & 5.96              & 5.27                         & 7.96         & 7.12                          & \textbf{5.15}                  \\
                             & BERT-small        & 29M                 & Generation        & 0.91          & 7.04            & 5.82              & 4.64                         & 4.89         & 8.78                          & 5.35                  \\
                             & BERT-medium       & 42M                 & Generation        & 0.93          & 5.73            & 5.57              & 5.07                         & 5.02         & 8.09                          & 5.07                  \\ \bottomrule
\end{tabular}
}
\caption{Full experimental results of measuring social bias in text generation metrics. PA: Physical Appearance. SS: Socioeconomic Status. The recommended (default) configurations are in \textbf{bold}.}
\label{tab:full_fairness_results}
\end{table*}

\section{Hyper-Parameters}
\label{sec:append_hyper}
We list our hyper-parameters for training debiasing adapters in Table~\ref{tab:hyper}. 
The hyper-parameters are tuned manually in a lightweight manner.
All experiments are conducted on a single NVIDIA 3090 GPU.

\begin{table}[h!]
    \centering
    \begin{tabular}{lccc}
    \toprule
     \textbf{Metric} & \textbf{LR} & \textbf{BSZ} & \textbf{Steps} \\ \midrule 
     BERTScore-base & 1e-4 & 32 & 150K\\
     BERTScore-large & 1e-4 & 16 & 300k\\
     BARTScore-base & 1e-3 & 32 & 100k\\
     BLEURT-base & 5e-4 & 16 & 300k\\\bottomrule
    \end{tabular}
    \caption{Hyper-parameters for training debiasing adapters. LR: learning rate. BSZ: batch size.}
    \label{tab:hyper}
\end{table}

\section{Full Results of Fairness Evaluation}
\label{sec:add_full_res}
We provide full results of evaluating metric bias in Table~\ref{tab:full_fairness_results}. For PLM-based metrics, we evaluate using different backbone models with varying sizes. For generation-based metrics, namely PRISM and BARTScore, we report the results of using precision, recall, and F-score as the text generation metric, respectively.

\section{Full Results of Performance Evaluation}
\label{sec:add_full_perf}
In Table~\ref{tab:miti_intrinsic}, we only show the average Pearson correlation of BERTScore and MoverScore across 10 language-pairs in the WMT20 dataset. Table~\ref{tab:full_pref_results} provides the full results of performance on all the language-pairs.

\begin{table*}[t]
\resizebox{\linewidth}{!}{
\begin{tabular}{lccccccccccc}
\toprule
\textbf{}    & \textbf{cs-en} & \textbf{de-en} & \textbf{iu-en} & \textbf{ja-en} & \textbf{km-en} & \textbf{pl-en} & \textbf{ps-en} & \textbf{ru-en} & \textbf{ta-en} & \textbf{zh-en} & \textbf{Avg.} \\ \midrule
\multicolumn{12}{l}{\textsc{BERTScore}}                                                                                                                                                                        \\ \midrule
BERT-large   & 0.733          & 0.803          & 0.631          & 0.865          & 0.979          & 0.401          & 0.937          & 0.861          & 0.820           & 0.929          & 0.796       \\
\ \ + Adapter & 0.738 & 0.792 & 0.639 & 0.866 & 0.976 & 0.367 & 0.936 & 0.856 & 0.823 & 0.927 & 0.792\\
Zari-Dropout & 0.798          & 0.799          & 0.661          & 0.815          & 0.942          & 0.421          & 0.919          & 0.878          & 0.820           & 0.914          & 0.797       \\
Zari-CDA     & 0.786          & 0.795          & 0.637          & 0.901          & 0.976          & 0.289          & 0.929          & 0.871          & 0.824          & 0.929          & 0.794       \\ 
BERT-base & 0.746 & 0.793 & 0.663 & 0.882 & 0.971 & 0.356 & 0.928 & 0.858 & 0.833 & 0.929 & 0.796 \\
\ \ + Adapter & 0.758 & 0.786 & 0.639 & 0.873 & 0.970 & 0.364 & 0.932 & 0.862 & 0.832 & 0.925 & 0.794 \\ \midrule
\multicolumn{12}{l}{\textsc{MoverScore}}                                                                                                                                                                       \\ \midrule
BERT-large   &  0.755 & 0.802 & 0.422 & 0.888 & 0.991 & 0.471 & 0.945 & 0.860  & 0.825 & 0.929 & 0.789 \\
Zari-Dropout &  0.812 & 0.788 & 0.433 & 0.876 & 0.985 & 0.442 & 0.917 & 0.859 & 0.840  & 0.928 & 0.788  \\
Zari-CDA     &  0.795 & 0.789 & 0.393 & 0.925 & 0.985 & 0.329 & 0.930  & 0.858 & 0.835 & 0.931 & 0.777  \\ \midrule
\multicolumn{12}{l}{\textsc{BLEURT}}                                                                                                                                                                       \\ \midrule
BERT-base & 0.754 & 0.832 & 0.486 & 0.806 & 0.976 & 0.317 & 0.956 & 0.838 & 0.779 & 0.918 & 0.766 \\
\ \ + Adapter & 0.780 & 0.758 & 0.605 & 0.873 & 0.996 & 0.493 & 0.976 & 0.884 & 0.789 & 0.916 & 0.807\\ \midrule
\multicolumn{12}{l}{\textsc{BARTScore}}                                                                                                                                                                       \\ \midrule
BART-base & 0.815 & 0.808 & 0.601 & 0.808 & 0.936 & 0.256 & 0.935 & 0.860 & 0.787 & 0.944 & 0.775 \\
\ \ + Adapter & 0.835 & 0.796 & 0.564 & 0.803 & 0.935 & 0.243 & 0.932 & 0.858 & 0.760 & 0.940 & 0.767\\
\ \ w/ Precision & 0.755 & 0.799 & 0.540 & 0.645 & 0.889 & 0.222 & 0.941 & 0.821 & 0.704 & 0.918 & 0.723 \\
\ \ w/ Recall & 0.747 & 0.682 & 0.642 & 0.731 & 0.970 & 0.191 & 0.829 & 0.861 & 0.664 & 0.941 & 0.726 \\
BART-large & 0.771 & 0.805 & 0.536 & 0.776 & 0.950 & 0.270 & 0.969 & 0.838 & 0.779 & 0.937 & 0.763 \\
\ \ w/ Precision & 0.721 & 0.809 & 0.474 & 0.604 & 0.898 & 0.233 & 0.958 & 0.790 & 0.721 & 0.919 & 0.713\\
\ \ w/ Recall & 0.749 & 0.569 & 0.575 & 0.828 & 0.986 & 0.229 & 0.947 & 0.831 & 0.800 & 0.924 & 0.744\\
\bottomrule
\end{tabular}
}
\caption{Full Pearson correlations of evaluated PLM-based metrics on WMT20 dataset.}
\label{tab:full_pref_results}
\end{table*}

\section{On the Definition of Metric Bias}
\label{sec:discussion}
In Eq. (\ref{eq:bias}) we measure the metric bias as the \textit{absolute difference} between the sentence pairs instead of the difference with the polarity of stereotype or anti-stereotype, which we will refer to as \textit{stereotypical difference}.

\paragraph{Why we use absolute difference?}
On the one hand, we adopt the absolute difference as the measurement of fairness because our purpose is to encourage text generation metrics to assign the same score to a pair of candidates if the only difference between them is the identity words instead of rating the stereotypical or anti-stereotypical one. If we use the stereotypical difference as the measurement of fairness, then a text generation metric that rates stereotypical candidates 50\% of the time and rates anti-stereotypical candidates 50\% of the time will be considered to be fair but actually, unfairness has happened to those candidates. We do not consider such a text generation metric to be fair though it seems fair "statistically". 

\paragraph{Results of stereotypical difference.}
On the other hand, stereotypical difference can be another useful measurement and is a good complementary to the current measurement. To that end, we also demonstrate results of gender bias evaluated using stereotypical difference in Table~\ref{tab:stereotypical_difference}. We find that both $n$-gram-based metrics and PLM-based metrics generally exhibit lower gender bias when switching to stereotypical difference but PLM-based metrics still carry a higher degree of gender bias than $n$-gram-based metrics. We leave the exploration of better measurement of metric bias to future work.

\begin{table}[t]
\centering
\resizebox{\linewidth}{!}{
\begin{tabular}{lcc}
\toprule
\textbf{Metric}           & \textbf{Absolute Diff.}         & \textbf{Stereotypical Diff.}       \\ \midrule
\multicolumn{3}{l}{$n$-gram-based metrics} \\ \midrule
BLEU             & 0.10       & 0.10     \\
ROUGE            & 0.21       & 0.21     \\
METEOR           & 1.08       & 0.11     \\
NIST             & 0.11       & 0.11     \\
chrF             & 1.23       & 0.15     \\ \midrule
\multicolumn{3}{l}{PLM-based metrics}    \\ \midrule
BERTScore        & 6.99       & 4.43     \\
MoverScore       & 13.24      & 1.67     \\
BLEURT           & 27.08      & 7.92     \\
PRISM            & 7.13       & 1.31     \\
BARTScore        & 9.47       & 3.54     \\
\bottomrule
\end{tabular}
}
\caption{Comparison of gender bias evaluated using absolute difference and stereotypical difference.}
\label{tab:stereotypical_difference}
\end{table}

\end{document}